\pdfoutput=1

\documentclass[11pt]{article}

\usepackage{acl}
\usepackage{times}
\usepackage{latexsym}
\usepackage{adjustbox}
\usepackage{xspace}
\usepackage{todonotes}
\usepackage{xcolor}
\usepackage{comment}
\usepackage{tikz}
\usepackage{pgfplots}
\usepackage{pgfplotstable}
\usepackage{makecell}
\usepackage{adjustbox}
\usepackage{pifont}
\usepackage{float}
\usepackage[edges]{forest}
\definecolor{hidden-draw}{RGB}{0,0,0}

\usepackage[T1]{fontenc}

\usepackage[utf8]{inputenc}

\usepackage{microtype}

\usepackage{inconsolata}

\usepackage{hyperref}
\usepackage[all]{hypcap}

\usepackage{tcolorbox}
\usepackage{amsmath}
\usepackage{amssymb}

\usepackage{booktabs}
\usepackage{threeparttable}
\usepackage{tabularx}

\usepackage{soul}
\usepackage{xcolor}
\usepackage{subfig}

\usepackage{bm} %

\usepackage{algorithm}
\usepackage{algorithmic}

\usepackage{wrapfig}

\newcommand{\ie}{{\em i.e.}}
\newcommand{\eg}{{\em e.g.}}
\newcommand{\aka}{{\em a.k.a.}}
\newcommand{\wrt}{{\em w.r.t.}}
\newcommand{\etc}{{\em inter alia}}

\definecolor{mygreen}{RGB}{11,141,10}
\definecolor{myred}{RGB}{223,68,52}
\definecolor{myblue}{RGB}{70,130,180}
\definecolor{mydeepblue}{RGB}{65,105,225}
\definecolor{myviolet}{RGB}{97,0,138}
\definecolor{myburgundy}{RGB}{110,10,30}
\definecolor{myblue2}{RGB}{0,105,148}
\definecolor{iceblue}{RGB}{173, 216, 230}
\definecolor{puregreen}{RGB}{0, 70, 0}

\definecolor{wingreen}{rgb}{0,0.45,0.24}
\definecolor{losered}{rgb}{1.0,0.1,0.24}

\definecolor{lightcoral}{rgb}{0.97, 0.36, 0.46}
\definecolor{lightyellow}{rgb}{0.98, 0.7, 0}
\definecolor{harvestgold}{rgb}{0.85, 0.57, 0.0}
\definecolor{brightlavender}{rgb}{0.75, 0.58, 0.89}
\definecolor{capri}{rgb}{0.0, 0.75, 1.0}
\definecolor{carminepink}{rgb}{0.92, 0.3, 0.26}
\definecolor{celadon}{rgb}{0.67, 0.88, 0.69}
\definecolor{darkpastelgreen}{rgb}{0.01, 0.75, 0.24}

\definecolor{grayhighlight}{RGB}{250,250,227}

\definecolor{target}{HTML}{F47983}
\definecolor{control}{HTML}{3E87CD}
\definecolor{credibility}{HTML}{B98AC9}
\definecolor{logical}{HTML}{93C572}
\definecolor{emotional}{HTML}{F9EAC3}

\newcommand{\cm}{\textbf{\textcolor{lightcoral}{CM}}}
\newcommand{\ic}{\textbf{\textcolor{lightyellow}{IC}}}
\newcommand{\im}{\textbf{\textcolor{cyan}{IM}}}

\usepackage{pifont}

\newcommand{\prehoc}{\raisebox{-0.6ex}{\includegraphics[scale=0.012]{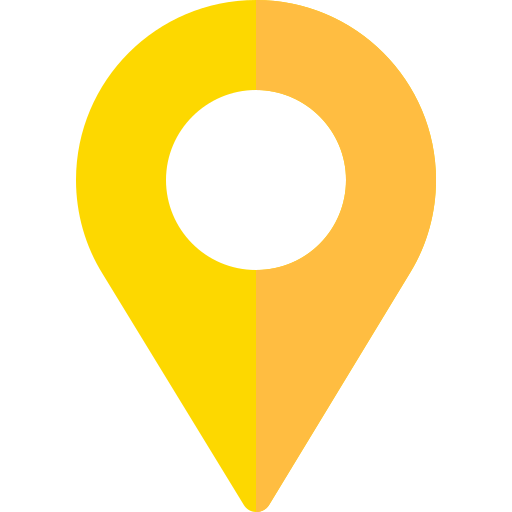}}}
\newcommand{\posthoc}{\raisebox{-0.6ex}{\includegraphics[scale=0.012]{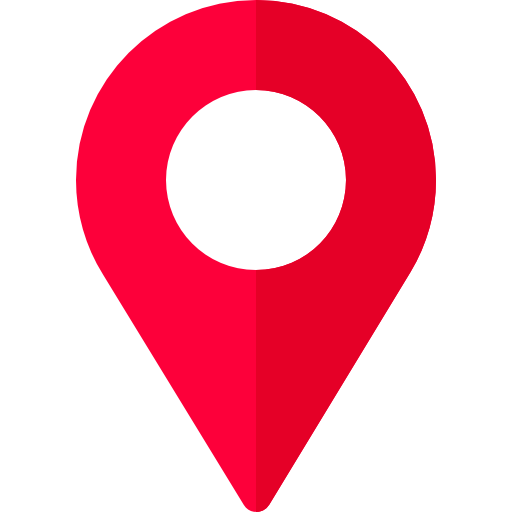}}}

\title{\raisebox{-0.8ex}{\includegraphics[scale=0.16]{Figs/Emoji/clash.pdf}} Knowledge Conflicts for LLMs: A Survey}

\author{Rongwu Xu$^{1\dag*}$, Zehan Qi$^{1\dag*}$, Zhijiang Guo$^{2\dag}$, \\ \bf Cunxiang Wang$^{3}$, Hongru Wang$^{4}$, Yue Zhang$^{3}$, Wei Xu$^{1}$ \\
$^{1}$Tsinghua University $^{2}$University of Cambridge \\$^{3}$Westlake University $^{4}$The Chinese University of Hong Kong\\
\texttt{\{xrw22, qzh23\}@mails.tsinghua.edu.cn} \\
$^{\dag}$ Leading authors, $^{*}$ Equal contribution
}

\begin{document}
\maketitle

\begin{abstract}
This survey provides an in-depth analysis of knowledge conflicts for large language models (LLMs), highlighting the complex challenges they encounter when blending contextual and parametric knowledge. 
Our focus is on three categories of knowledge conflicts: context-memory, inter-context, and intra-memory conflict. These conflicts can significantly impact the trustworthiness and performance of LLMs, especially in real-world applications where noise and misinformation are common. 
By categorizing these conflicts, exploring the causes, examining the behaviors of LLMs under such conflicts, and reviewing available solutions, this survey aims to shed light on strategies for improving the robustness of LLMs, thereby serving as a valuable resource for advancing research in this evolving area.
\end{abstract}

\noindent
\begin{wrapfigure}{l}{0.05\textwidth}
    \centering
    \hypertarget{github-link}{}
    \href{https://github.com/pillowsofwind/Knowledge-Conflicts-Survey}{%
    \includegraphics[width=0.05\textwidth]{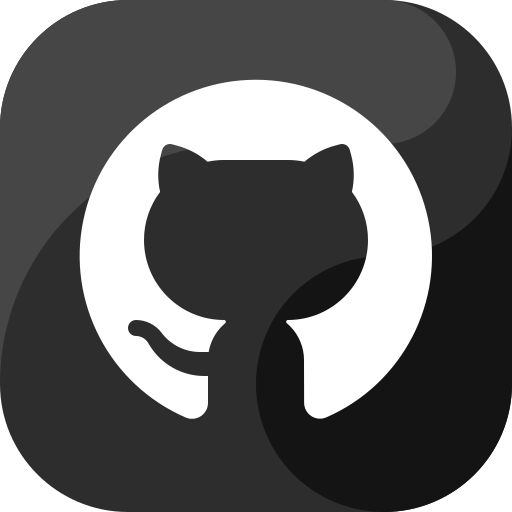}
    }
\vspace{-2em}
\end{wrapfigure}

\noindent
{\fontsize{10}{10}\selectfont\url{https://github.com/pillowsofwind/Knowledge-Conflicts-Survey}}

\section{Introduction}
\label{sec:intro}

Large language models (LLMs;~\citealt{brown2020language,touvron2023llama, openai2023gpt4}) are renowned for encapsulating a vast repository of world knowledge~\cite{roberts2020much,HuFacts2023}, referred to as \emph{parametric knowledge}. These models excel in knowledge-intensive tasks including QA \cite{petroni2019language}, fact-checking \cite{gao2023rarr}, dialogue system \cite{wang2023dialoguesurvey}, knowledge generation~\cite{chen2023beyond}, \etc. 
In the meantime, LLMs continue to engage with external \emph{contextual knowledge} after deployed~\cite{pan2022knowledge}, including user prompts~\cite{liu2023pre}, interactive dialogues~\cite{zhang2020dialogpt, wang2024unimsrag}, or retrieved documents from the Web~\cite{lewis2020retrieval,shi2023replug}, and tools~\cite{schick2023toolformer,zhuang2023toolqa}.

\begin{figure}
    \centering
    \includegraphics[width=\linewidth]{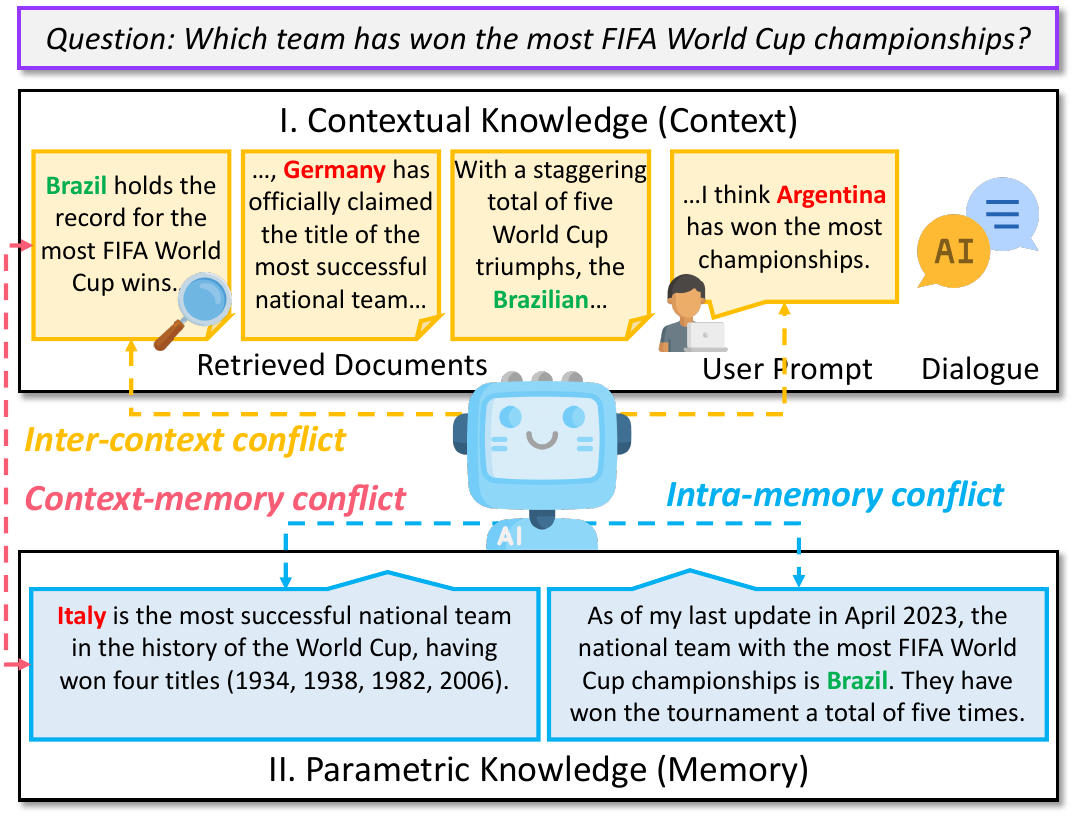}
    \caption{An LLM may encounter three distinct types of knowledge conflicts, stemming from knowledge sources—either contextual (\emph{I. Context}, \textcolor{orange!70}{yellow} chatboxes) or inherent to the LLM's parameters (\emph{II. Memory}, \textcolor{cyan}{blue} chatboxes). When confronted with a user's question (\textcolor{myviolet}{purple} chatbox) entailing knowledge of complex conflicts, the LLM is required to resolve these discrepancies to deliver accurate responses.}
    \label{fig:overview}
\end{figure}

\begin{figure*}
    \centering
    \includegraphics[width=0.8\linewidth]{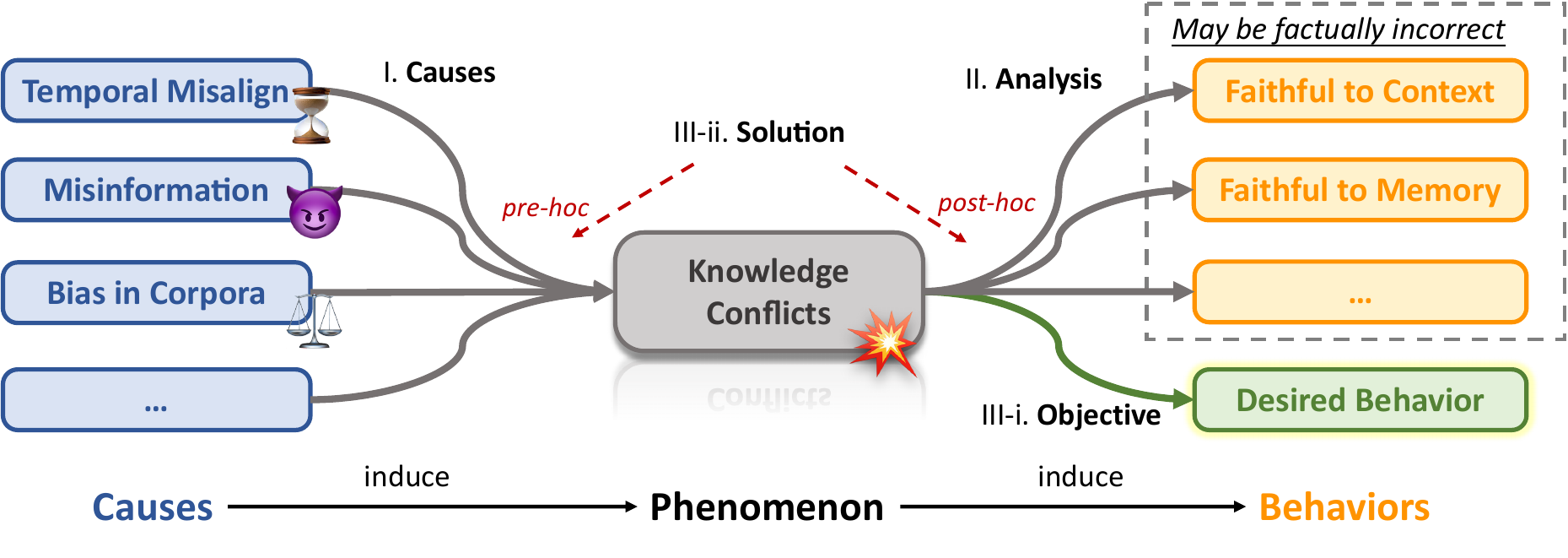}
    \caption{We view knowledge conflict not only as a standalone \textbf{phenomenon} but also as a nexus that connects various causal triggers (\textbf{causes}) with the \textbf{behaviors} of LLMs. While existing literature mainly focuses on \emph{II. Analysis}, our survey involves systematically observing these conflicts, offering insights into their emergence and impact on LLMs' behavior, along with the desirable behaviors and related solutions.}
    \label{fig:methodology}
\end{figure*}

Integrating contextual knowledge into LLMs enables them to keep abreast of current events \cite{kasai2022realtime} and generate more accurate responses \cite{shuster2021retrieval}, yet it risks conflicting due to the rich knowledge sources. 
The discrepancies \emph{among} the contexts and the model's parametric knowledge are referred to as \emph{knowledge conflicts}~\cite{chen2022rich,xie2023adaptive}.
In this paper, we categorize three distinct types of knowledge conflicts, as characterized in~\autoref{fig:overview}. 
As the example shown in~\autoref{fig:overview}, when utilizing an LLM to respond to a user question, users may provide the LLM with supplementary prompts, while the LLM also leverages search engines to gather relevant documents from the Web to enhance its knowledge~\cite{lewis2020retrieval}. This combination of user prompts, dialogue history, and retrieved documents constitutes contextual knowledge (\emph{context}). Contextual knowledge can conflict with the parametric knowledge (\emph{memory}) encapsulated within the LLM's parameters \citep{longpre2021entity, xie2023adaptive}, a phenomenon we term as \textbf{context-memory conflict} (\cm, \autoref{sec:context-memory}).
In real-world scenarios, the external document might be fraught with noise \cite{zhang2021situatedqa} or even deliberately crafted misinformation \cite{du2022synthetic,pan2023attacking}, complicating their ability to process and respond accurately~\cite{chen2022rich}.
We term the conflict among various pieces of contextual knowledge as \textbf{inter-context conflict} (\ic, \autoref{sec:inter-context}).
To reduce uncertainties in responses, the user may pose the question in various forms. Therefore, the LLM's parametric knowledge may yield divergent responses to these differently phrased questions. This variance can be attributed to the conflicting knowledge embedded within the LLM's parameters, which stem from the inconsistencies present in the complex and diverse pre-training data sets~\cite{huang2023survey}. This gives rise to what we term as \textbf{intra-memory conflict} (\im, \autoref{sec:intra-memory}).

Knowledge conflict is originally rooted in open-domain QA research. The concept gained attention in~\citet{longpre2021entity} that focused on the entity-based conflicts between parametric knowledge and external passages. Concurrently, discrepancies among multiple passages were also scrutinized subsequently~\cite{chen2022rich}.
Knowledge conflicts attract significant attention with the recent advent of LLMs.
For instance, recent studies find that %
LLMs exhibit both adherence to parametric knowledge and susceptibility to contextual influences~\cite{xie2023adaptive}, which can be problematic when this external knowledge is factually incorrect~\cite{pan2023risk}.
Given the implications for the trustworthiness \citep{du2022synthetic}, real-time accuracy \citep{kasai2022realtime}, and robustness of LLMs \citep{ying2023intuitive}, it is imperative to delve deeper into understanding and resolving knowledge conflicts \citep{xie2023adaptive,wang2023resolving}.

As of the time of writing, to the best of our knowledge, there is no systematic survey dedicated to the investigation of knowledge conflicts. 
Existing reviews~\cite{zhang2023sirens, wang2023survey, feng2023trends} touch upon knowledge conflicts as a subtopic within their broader contexts.
While \citet{feng2023trends} offer a more systematic examination of knowledge conflicts, categorizing them into external and internal conflicts. However, their survey provides only a brief overview of relevant works and primarily focuses on specific scenarios.
To fill the gap, we aim to provide a comprehensive review encompassing the categorization, cause and behavior analysis, and solutions for addressing various kinds of knowledge conflicts. 

We conceptualize the \emph{lifecycle of knowledge conflicts} as both a \emph{cause} leading to various behaviors, and an \emph{effect} emerges from the intricate nature of knowledge as in \autoref{fig:methodology}. Knowledge conflicts serve as a crucial intermediary between causes and model behaviors. For instance, they significantly contribute to the model generating factually incorrect information, \aka, hallucinations~\cite{ji2023survey, zhang2023sirens}.
Our research, in a manner akin to Freudian psychoanalysis, underscores the significance of understanding the origins of these conflicts. Although existing analyses \cite{chen2022rich,xie2023adaptive, wang2023resolving} tend to construct such conflicts artificially, we posit that these analyses do not sufficiently address the interconnectedness of the issue.

\begin{figure*}[t!]
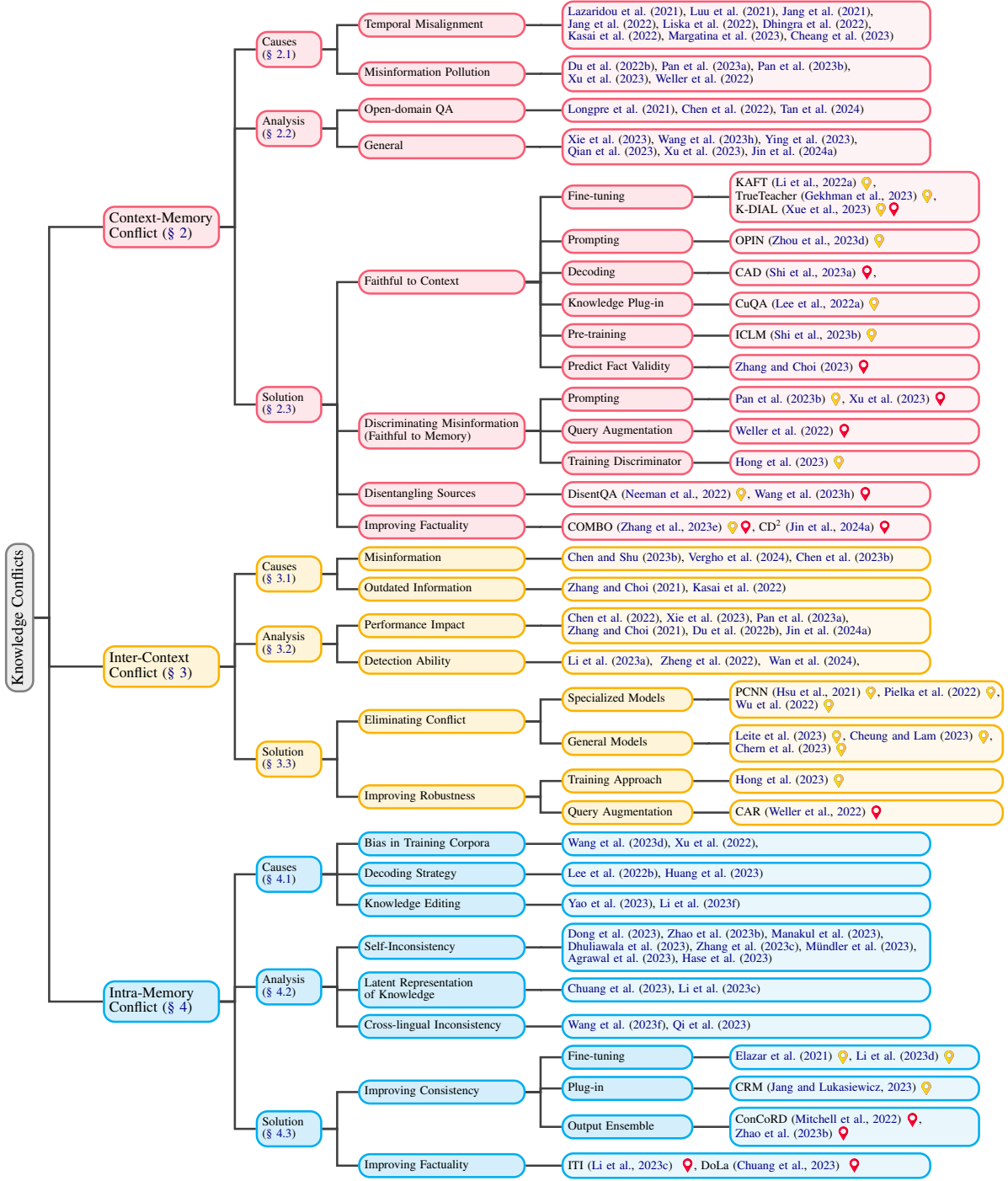

    \centering
    \tikzstyle{my-box}=[
    rectangle,
    rounded corners,
    text opacity=1,
    minimum height=0.1em,
    minimum width=0.1em,
    inner sep=2pt,
    align=left,
    fill opacity=.5,
    ]
    \tikzstyle{leaf}=[my-box]
    \resizebox{0.99\textwidth}{!}{
        \begin{forest}
            forked edges,
            for tree={
                grow=east,
                reversed=true,
                anchor=base west,
                parent anchor=east,
                child anchor=west,
                base=left,
                font=\fontsize{6}{6}\selectfont,
                rectangle,
                draw=hidden-draw,
                rounded corners,
                align=left,
                minimum width=0.1em,
                edge+={darkgray, line width=0.8pt},
                s sep=2pt,
                inner xsep=2pt,
                inner ysep=2pt,
                ver/.style={rotate=90, child anchor=north, parent anchor=south, anchor=center},
            },
            where level=1{text width=3.5em,font=\fontsize{5.5}{5.5}\selectfont,}{},
            where level=2{text width=1.8em,font=\fontsize{4.5}{4.5}\selectfont,}{},
            where level=3{text width=5.2em,font=\fontsize{4.5}{4.5}\selectfont,}{},
            where level=4{text width=4em,font=\fontsize{4.5}{4.5}\selectfont,}{},
            where level=5{text width=8em,font=\fontsize{4.5}{4.5}\selectfont,}{},
            [
                Knowledge Conflicts, draw=gray, color=gray!100, fill=gray!15, thick, text=black, ver
                [
                    Context-Memory \\ Conflict (\autoref{sec:context-memory})
                    , color=lightcoral!100, fill=lightcoral!15, thick, text=black
                    [
                        Causes \\(\autoref{subsec:cm-causes})
                        , color=lightcoral!100, fill=lightcoral!15, thick, text=black
                        [
                            Temporal Misalignment
                            , color=lightcoral!100, fill=lightcoral!15, thick, text=black
                            [   
                                \citet{lazaridou2021mind}{,}
                                \citet{luu2021time}{,}
                                \citet{jang2021towards}{,}\\
                                \citet{jang2022temporalwiki}{,}
                                \citet{liska2022streamingqa}{,}
                                \citet{dhingra2022time}{,}\\
                                \citet{kasai2022realtime}{,}
                                \citet{margatina2023dynamic}{,}
                                \citet{cheang2023can}
                                , leaf, text width = 12em, color=lightcoral!100, fill=lightcoral!15, thick, text=black
                            ]
                        ]
                        [
                            Misinformation Pollution
                            , color=lightcoral!100, fill=lightcoral!15, thick, text=black
                            [   
                                \citet{du2022synthetic}{,}
                                \citet{pan2023attacking}{,}
                                \citet{pan2023risk}{,}\\ 
                                \citet{xu2023earth}{,} \citet{weller2022defending}
                                , leaf, text width = 12em, color=lightcoral!100, fill=lightcoral!15, thick, text=black
                            ]
                        ]
                    ]
                    [
                        Analysis \\(\autoref{subsec:cm-analysis})
                        , color=lightcoral!100, fill=lightcoral!15, thick, text=black
                        [
                            Open-domain QA
                            , color=lightcoral!100, fill=lightcoral!15, thick, text=black
                            [
                                \citet{longpre2021entity}{,}
                                \citet{chen2022rich}{,}
                                \citet{tan2024blinded}
                                , leaf, text width = 12em, color=lightcoral!100, fill=lightcoral!15, thick, text=black
                            ]
                        ]
                        [
                            General
                            , color=lightcoral!100, fill=lightcoral!15, thick, text=black
                            [
                                \citet{xie2023adaptive}{,} 
                                \citet{wang2023resolving}{,} 
                                \citet{ying2023intuitive}{,}\\
                                \citet{qian2023merge}{,}
                                \citet{xu2023earth}{,}
                                \citet{jin2024tug}  
                                , leaf, text width = 12em, color=lightcoral!100, fill=lightcoral!15, thick, text=black
                            ]
                        ]
                    ]
                    [
                        Solution \\(\autoref{subsec:cm-mitigation})
                        , color=lightcoral!100, fill=lightcoral!15, thick, text=black
                        [   
                            Faithful to Context
                            , color=lightcoral!100, fill=lightcoral!15, thick, text=black
                            [
                                Fine-tuning
                                , color=lightcoral!100, fill=lightcoral!15, thick, text=black
                                [
                                    KAFT~\citep{li2022large}{ \prehoc}{,}\\
                                    TrueTeacher~\citep{gekhman2023trueteacher}{ \prehoc}{,}\\
                                    K-DIAL~\citep{xue2023improving}{ \prehoc \posthoc}
                                    , leaf, color=lightcoral!100, fill=lightcoral!15, thick, text=black
                                ]
                            ]
                            [
                                Prompting
                                , color=lightcoral!100, fill=lightcoral!15, thick, text=black
                                [
                                    OPIN~\citep{zhou2023context}{ \prehoc}
                                     , leaf, color=lightcoral!100, fill=lightcoral!15, thick, text=black
                                ]
                            ]
                            [
                                Decoding
                                , color=lightcoral!100, fill=lightcoral!15, thick, text=black
                                [
                                    CAD~\citep{shi2023trusting}{ \posthoc}{,}\\
                                    , leaf, color=lightcoral!100, fill=lightcoral!15, thick, text=black
                                ]
                            ]
                            [
                                Knowledge Plug-in
                                , color=lightcoral!100, fill=lightcoral!15, thick, text=black
                                [
                                    CuQA~\citep{lee2022plug}{ \prehoc}
                                    , leaf, color=lightcoral!100, fill=lightcoral!15, thick, text=black
                                ]
                            ]
                            [
                                Pre-training
                                , color=lightcoral!100, fill=lightcoral!15, thick, text=black
                                [
                                    ICLM~\citep{shi2023context}{ \prehoc}
                                    , leaf, color=lightcoral!100, fill=lightcoral!15, thick, text=black
                                ]
                            ]
                            [
                                Predict Fact Validity
                                , color=lightcoral!100, fill=lightcoral!15, thick, text=black
                                [
                                    \citet{zhang2023mitigating}{ \posthoc}
                                    , leaf, color=lightcoral!100, fill=lightcoral!15, thick, text=black
                                ]
                            ]
                        ]
                        [
                            Discriminating Misinformation\\(Faithful to Memory)
                            , color=lightcoral!100, fill=lightcoral!15, thick, text=black   
                                [
                                    Prompting
                                    , color=lightcoral!100, fill=lightcoral!15, thick, text=black
                                    [
                                        \citet{pan2023risk}{ \prehoc{}}{,}
                                        \citet{xu2023earth}{ \posthoc}
                                        , leaf, color=lightcoral!100, fill=lightcoral!15, thick, text=black
                                    ]
                                ]
                                [
                                    Query Augmentation
                                    ,
                                    color=lightcoral!100, fill=lightcoral!15, thick, text=black
                                    [
                                        \citet{weller2022defending}{ \posthoc}
                                        , leaf, color=lightcoral!100, fill=lightcoral!15, thick, text=black
                                    ]
                                ]
                                [
                                    Training Discriminator
                                    ,
                                    color=lightcoral!100, fill=lightcoral!15, thick, text=black
                                    [
                                        \citet{hong2023discern}{ \prehoc}
                                        , leaf, color=lightcoral!100, fill=lightcoral!15, thick, text=black
                                    ]
                                ]
                        ]
                        [
                            Disentangling Sources
                            ,
                            color=lightcoral!100, fill=lightcoral!15, thick, text=black
                            [
                                        DisentQA~\citep{neeman2022disentqa}{ \prehoc}{,}
                                        \citet{wang2023resolving}{ \posthoc}
                                        , leaf, text width=12em, color=lightcoral!100, fill=lightcoral!15, thick, text=black
                            ]
                        ]
                        [
                            Improving Factuality
                            ,
                            color=lightcoral!100, fill=lightcoral!15, thick, text=black
                            [
                                        COMBO~\citep{zhang2023merging}{ \prehoc{}\posthoc}{,}
                                        CD$^{\text{2}}$~\citep{jin2024tug}{ \posthoc}
                                        , leaf, text width=12em, color=lightcoral!100, fill=lightcoral!15, thick, text=black
                            ]
                        ]
                    ]
                ]
                [
                    Inter-Context\\ Conflict (\autoref{sec:inter-context})
                    , color=lightyellow!100, fill=lightyellow!15, thick, text=black
                    [
                        Causes\\(\autoref{subsec:ic-causes})
                    , color=lightyellow!100, fill=lightyellow!15, thick, text=black
                        [
                           Misinformation, color=lightyellow!100, fill=lightyellow!15, thick, text=black
                           [
                                \citet{chen2023combating}{,} \citet{vergho2024comparing}{,}
                                \citet{chen2023benchmarking}
                                , leaf, color=lightyellow!100, fill=lightyellow!15, thick, text=black, text width=12em
                           ]
                        ]
                        [
                            Outdated Information, color=lightyellow!100, fill=lightyellow!15, thick, text=black
                            [
                                \citet{zhang2021situatedqa}{,}
                                \citet{kasai2022realtime}, leaf, color=lightyellow!100, fill=lightyellow!15, thick, text=black, text width=12em
                            ]
                        ]
                    ]
                    [
                        Analysis \\(\autoref{subsec:ic-analysis})
                        , color=lightyellow!100, fill=lightyellow!15, thick, text=black, 
                        [   
                            Performance Impact
                            , color=lightyellow!100, fill=lightyellow!15, thick, text=black, 
                            [
                                \citet{chen2022rich}{,} 
                                \citet{xie2023adaptive}{,} 
                                \citet{pan2023attacking}{,} \\
                                \citet{zhang2021situatedqa}{,}
                                \citet{du2022synthetic}{,} 
                                \citet{jin2024tug}, leaf, color=lightyellow!100, fill=lightyellow!15, thick, text=black, text width=12em
                            ]
                        ]
                        [
                            Detection Ability
                            , color=lightyellow!100, fill=lightyellow!15, thick, text=black, 
                            [
                                \citet{li2023contradoc}{, }
                                \citet{zheng2022cdconv}{, }
                                \citet{wan2024evidence}{, }
                                , leaf, color=lightyellow!100, fill=lightyellow!15, thick, text=black, text width=12em
                            ]
                        ]
                    ]
                    [
                        Solution \\(\autoref{subsec:ic-mitigation})
                        , color=lightyellow!100, fill=lightyellow!15, thick, text=black
                        [
                            Eliminating Conflict 
                            , color=lightyellow!100, fill=lightyellow!15, thick, text=black
                            [
                                Specialized Models
                                , color=lightyellow!100, fill=lightyellow!15, thick, text=black
                                [
                                    PCNN \citep{hsu2021wikicontradiction} {\prehoc}{,}
                                    \citet{pielka2022linguistic}{ \prehoc}{,}\\
                                    \citet{wu2022topological}{ \prehoc}
                                    , leaf, color=lightyellow!100, fill=lightyellow!15, thick, text=black, text width=8.8em
                                ]
                            ]
                            [
                                General Models
                                , color=lightyellow!100, fill=lightyellow!15, thick, text=black
                                [
                                    \citet{leite2023detecting}{ \prehoc}{,}
                                    \citet{cheung2023factllama}{ \prehoc}{,} \\
                                    \citet{chern2023factool}{ \prehoc}
                                    , leaf, color=lightyellow!100, fill=lightyellow!15, thick, text=black, text width=8.8em
                                ]
                            ]
                        ]
                        [
                            Improving Robustness
                            , color=lightyellow!100, fill=lightyellow!15, thick, text=black
                            [
                                Training Approach, color=lightyellow!100, fill=lightyellow!15, thick, text=black
                                [
                                    \citet{hong2023discern}{ \prehoc} 
                                    , leaf, color=lightyellow!100, fill=lightyellow!15, thick, text=black, text width=8.8em
                                ]
                            ]
                            [
                                Query Augmentation, color=lightyellow!100, fill=lightyellow!15, thick, text=black
                                [
                                    CAR \citep{weller2022defending}{ \posthoc}
                                    , leaf, color=lightyellow!100, fill=lightyellow!15, thick, text=black, text width=8.8em
                                ]
                            ]
                        ]
                    ]
                ]
                [
                    Intra-Memory \\Conflict (\autoref{sec:intra-memory})
                    , color=cyan!100, fill=cyan!15, thick, text=black
                    [
                        Causes \\(\autoref{subsec:im-causes}), color=cyan!100, fill=cyan!15, thick, text=black
                        [
                            Bias in Training Corpora, color=cyan!100, fill=cyan!15, thick, text=black
                            [
                                \citet{wang2023causal}{,} 
                                \citet{xu2022does}{,}
                                ,leaf,color=cyan!100, fill=cyan!15, thick, text=black, text width=12em
                            ]
                        ]
                        [
                            Decoding Strategy, color=cyan!100, fill=cyan!15, thick, text=black
                            [
                                \citet{lee2022factuality}{,} 
                                \citet{huang2023survey}, leaf,color=cyan!100, fill=cyan!15, thick, text=black, text width=12em
                            ]
                        ]
                        [
                            Knowledge Editing, color=cyan!100, fill=cyan!15, thick, text=black
                            [
                                \citet{yao2023editing}{,}
                                \citet{li2023unveiling}, leaf,color=cyan!100, fill=cyan!15, thick, text=black, text width=12em
                            ]
                        ]
                    ]
                    [
                        Analysis \\(\autoref{subsec:im-analysis})
                        , color=cyan!100, fill=cyan!15, thick, text=black
                        [
                            Self-Inconsistency
                            , color=cyan!100, fill=cyan!15, thick, text=black
                            [
                                \citet{dong2023statistical}{,}
                                \citet{zhao2023knowing}{,} 
                                \citet{manakul2023selfcheckgpt}{,} \\
                                \citet{dhuliawala2023chain}{,}
                                \citet{zhang2023sac}{,} 
                                \citet{mundler2023self}{,} \\
                                \citet{agrawal2023language}{,} 
                                \citet{hase2023methods}
                                , leaf,color=cyan!100, fill=cyan!15, thick, text=black, text width=12em
                            ]
                        ]
                        [
                            Latent Representation\\ of Knowledge
                            , color=cyan!100, fill=cyan!15, thick, text=black
                            [
                                \citet{chuang2023dola}{,}
                                \citet{li2023inference}
                                , leaf,color=cyan!100, fill=cyan!15, thick, text=black, text width=12em
                            ]
                        ]
                        [ 
                            Cross-lingual Inconsistency
                            , color=cyan!100, fill=cyan!15, thick, text=black
                            [
                                \citet{wang2023cross}{,}
                                \citet{qi2023cross}
                                ,leaf,color=cyan!100, fill=cyan!15, thick, text=black, text width=12em
                            ]
                        ]   
                    ]
                    [
                        Solution \\(\autoref{subsec:im-mitigation})
                        ,color=cyan!100, fill=cyan!15, thick, text=black
                        [
                            Improving Consistency
                            , color=cyan!100, fill=cyan!15, thick, text=black
                            [
                                Fine-tuning
                                , color=cyan!100, fill=cyan!15, thick, text=black
                                [
                                    \citet{elazar2021measuring}{ \prehoc}{,} 
                                    \citet{li2023benchmarking}{ \prehoc}, leaf, color=cyan!100, fill=cyan!15, thick, text=black, text width=8em
                                ]
                            ]
                            [
                                Plug-in
                                ,color=cyan!100, fill=cyan!15, thick, text=black
                                [
                                    CRM \citep{jang2023improving}{ \prehoc}
                                    ,leaf, color=cyan!100, fill=cyan!15, thick, text=black, text width=8em
                                ]
                            ]
                            [
                                Output Ensemble
                                ,color=cyan!100, fill=cyan!15, thick, text=black
                                [
                                    ConCoRD \citep{mitchell2022enhancing}{ \posthoc}{,}\\
                                    \citet{zhao2023knowing}{ \posthoc}
                                    ,leaf, color=cyan!100, fill=cyan!15, thick, text=black, text width=8em
                                ]
                            ]
                        ]
                        [
                            Improving Factuality
                            , color=cyan!100, fill=cyan!15, thick, text=black
                            [
                                ITI \citep{li2023inference} { \posthoc}{,}
                                DoLa \citep{chuang2023dola} { \posthoc},leaf, color=cyan!100, fill=cyan!15, thick, text=black, text width=12em
                            ]
                        ]
                    ]
                ]
            ]
        \end{forest}
    }
    \caption{Taxonomy of knowledge conflicts. We mainly list works in the era of LLMs. \includegraphics[scale=0.02]{Figs/Emoji/yellowpin.png} denotes pre-hoc solution and \includegraphics[scale=0.02]{Figs/Emoji/redpin.png} denotes post-hoc solution.}
    \label{fig:taxonomy}
\end{figure*}

Going beyond reviewing and analyzing causes and behaviors, we delve deeper to provide a systematic review of solutions, which are employed to minimize the undesirable consequences of knowledge conflicts, \ie, to encourage the model to exhibit \emph{desired behaviors that conform to specific objectives} (please noted that these \emph{objectives may differ} based on the particular scenario).
Based on the timing relative to potential conflicts, strategies are divided into two categories: \emph{pre-hoc} and \emph{post-hoc} strategies. The key distinction between them lies in whether adjustments are made \emph{before} or \emph{after} potential conflicts arise\footnote{Another interpretation is that a pre-hoc strategy is proactive while a post-hoc one is reactive.}.
The taxonomy of knowledge conflicts is outlined in~\autoref{fig:taxonomy}. We sequentially discuss the three kinds of knowledge conflicts, detailing for each the causes, analysis of model behaviors, and available solutions organized according to their respective objectives.
Related datasets can be found in~\autoref{tab:datasets}.

\begin{table*}[h]
\fontsize{10}{10}\selectfont
\centering
\setlength{\tabcolsep}{2pt} %
\begin{threeparttable}
\begin{tabularx}{\textwidth}{>{\centering\arraybackslash}p{3.35cm}>{\centering\arraybackslash}p{1.2cm}>{\centering\arraybackslash}X>{\centering\arraybackslash}p{0.9cm}>{\centering\arraybackslash}p{1.1cm}}
\toprule
\textbf{Datasets} & \textbf{Approach\tnote{1}} & \textbf{Base\tnote{2}} & \textbf{Size} & \textbf{Conflict}\\
\midrule
\citet{xie2023adaptive} & Gen &PopQA~(\citeyear{mallen2023not}), \textsc{StrategyQA}~(\cite{geva2021did}) & 20,091 & \cm{}\tnote{3} \\
KC~(\citeyear{wang2023resolving}) & Sub &N/A (LLM generated) & 9,803 & \cm{} \\
KRE~(\citeyear{ying2023intuitive}) & Gen &{\fontsize{9}{10}\selectfont MuSiQue~(\citeyear{trivedi2022musique}), SQuAD2.0~(\citeyear{rajpurkar2018know}), ECQA~(\citeyear{aggarwal2021explanations}), e-CARE~(\citeyear{du2022care})}  & 11,684 & \cm{} \\
Farm~(\citeyear{xu2023earth}) & Gen &BoolQ~(\citeyear{clark2019boolq}), NQ~(\citeyear{kwiatkowski2019natural}), TruthfulQA~(\citeyear{lin2022truthfulqa})  & 1,952 & \cm{} \\
\citet{tan2024blinded} & Gen &NQ~(\citeyear{kwiatkowski2019natural}), TriviaQA~(\citeyear{joshi2017triviaqa})  & 14,923 & \cm{} \\
WikiContradiction~(\citeyear{hsu2021wikicontradiction}) & Hum & Wikipedia & 2,210 & \ic{} \\
ClaimDiff~(\citeyear{ko2022claimdiff}) & Hum & N/A & 2,941 & \ic{} \\
\citet{pan2023attacking} & Gen,Sub &  SQuAD v1.1~(\citeyear{rajpurkar2016squad}) & 52,189 & \ic{} \\
\textsc{ContraDoc}~(\citeyear{li2023contradoc}) & Gen & CNN-DailyMail~(\citeyear{hermann2015teaching}), NarrativeQA~(\citeyear{kovcisky2018narrativeqa}), WikiText~(\citeyear{merity2016pointer}) & 449 & \ic{} \\
\textsc{ConflictingQA}~(\citeyear{wan2024evidence}) & Gen & N/A & 238 & \ic{} \\
\textsc{ParaRel}~(\citeyear{elazar2021measuring}) & Hum & T-REx~(\citeyear{elsahar2018t}) & 328 & \im{} \\
\bottomrule
\end{tabularx}
\end{threeparttable}
\fontsize{9.5}{9.5}\selectfont
\begin{tablenotes}
   \item{1.} Approach refers to how the conflicts are crafted, including entity-level substitution (Sub), generative approaches employing an LLM (Gen), and human annotation (Hum). 
   \item{2.} Base refers to the base dataset(s) that serve as the foundation for generating conflicts, if applicable.
   \item{3.} \includegraphics[scale=0.1]{Figs/Emoji/caution.pdf} 
   When using \cm{} datasets, conflicts arise from the specific model's parametric knowledge, which can differ across models. Therefore, selecting a subset of the dataset that aligns with the tested model's knowledge is crucial.
\end{tablenotes}
\caption{\label{tab:datasets}Datasets on evaluating LLMs' behavior when encountering knowledge conflicts. \cm{}: context-memory conflict, \ic{}: inter-context conflict, \im{}: intra-memory conflict.}
\end{table*}

\section{\textcolor{lightcoral}{C}ontext-\textcolor{lightcoral}{M}emory Conflict}
\label{sec:context-memory}

Context-memory conflict emerges as the most extensively investigated among the three types of conflicts.
LLMs are characterized by fixed parametric knowledge, a result of the substantial pertaining process~\cite{sharir2020cost,hoffmann2022training,craig2023large}. This static parametric knowledge stands in stark contrast to the dynamic nature of external information, which evolves at a rapid pace~\cite{de2021editing, kasai2022realtime}. 

\subsection{Causes}
\label{subsec:cm-causes}

The core of context-memory conflict stems from a discrepancy between the context and parametric knowledge.
We consider two main causes: temporal misalignment~\cite{lazaridou2021mind,luu2021time,dhingra2022time} and misinformation pollution~\cite{du2022synthetic,pan2023attacking}.

\noindent \textbf{Temporal Misalignment.}
Temporal misalignment \emph{naturally} arises in models trained on data collected in the past, as they may not accurately reflect contemporary or future realities (\ie, the contextual knowledge after the deployment)~\cite{luu2021time, lazaridou2021mind, liska2022streamingqa}. Such misalignment can degrade the model's performance and relevancy over time, as it may fail to capture new trends, shifts in language use, cultural changes, or updates in knowledge. Researchers have noted that temporal misalignment relegates the model's performance on various NLP tasks~\cite{luu2021time, zhang2021situatedqa, dhingra2022time, kasai2022realtime, cheang2023can}. 
Furthermore, the issue of temporal misalignment is expected to intensify due to the pre-training paradigm and the escalating costs associated with scaling up models~\citep{KaplanScaling20}.

Prior work tries to tackle temporal misalignment by focusing on three lines of strategies: \emph{Knowledge editing (KE)} aims to directly update the parametric knowledge of an existing pre-trained model~\citep{sinitsin2019editable, de2021editing, mitchell2021fast, onoe2023can}. \emph{Retrieval-augmented generation (RAG)} leverages a retrieval module to fetch relevant documents from external sources (\eg, database, the Web) to supplement the model's knowledge without altering its parameters~\cite{karpukhin2020dense, guu2020retrieval, lewis2020retrieval, lazaridou2022internet, borgeaud2022improving, peng2023check, vu2023freshllms}.
\emph{Continue learning (CL)} seeks to update the internal knowledge through continual pre-training on new and updated data~\cite{lazaridou2021mind, jang2021towards,jang2022temporalwiki}.
However, these methods of mitigating temporal misalignment are not magic bullets. KE can bring in side effects of knowledge conflict, leading to knowledge inconsistency (\ie, a sort of intra-memory conflict) and may even enhance the hallucination of LLMs~\cite{li2023unveiling, pinter2023emptying}. 
For RAG, it is inevitable to encounter knowledge conflicts since the model's parameters are not updated~\cite{chen2021dataset, zhang2021situatedqa}.
CL suffers from catastrophic forgetting issues and demands significant computational resources~\cite {de2021continual,he2021analyzing,wang2023incorporating}.

\noindent \textbf{Misinformation Pollution.}
Misinformation pollution emerges as another contributor to context-memory conflict, particularly for time-invariant knowledge~\cite{jang2021towards}. Adversaries exploit this vulnerability by introducing false or misleading information into the Web corpus of retrieved documents~\cite{pan2023attacking, pan2023risk, weller2022defending} and user conversations~\cite{xu2023earth, HuEvaluate24}. 
The latter poses a practical threat, as adversaries can leverage techniques such as \emph{prompt injection} attacks~\cite{liu2023prompt, greshake2023more, yi2023benchmarking}.
This vulnerability poses a real threat, as models might unknowingly propagate misinformation if they incorporate deceptive inputs without scrutiny~\cite{xie2023adaptive, pan2023risk, xu2023earth}.

Fabricated, malicious misinformation can markedly undermine the accuracy of automated fact-checking~\cite{du2022synthetic} and open-domain question-answering systems~\cite{pan2023attacking, pan2023risk}. Furthermore, recent studies also highlight the model's tendency to align with user opinions, \aka, \emph{sycophancy}, further exacerbating the issue~\cite{perez2022discovering, turpin2023language, wei2023simple, sharma2023towards}.
In the current landscape of LLMs, there is growing apprehension in the NLP community regarding the potential generation of misinformation by LLMs~\cite{ayoobi2023looming, kidd2023ai, carlini2023poisoning, zhou2023synthetic, spitale2023ai, chen2023combating}. Researchers acknowledge the challenges associated with detecting misinformation generated by LLMs~\cite{tang2023science, chen2023can, jiang2023disinformation}. This underscores the urgency of addressing the nuanced challenges posed by LLMs in the context of contextual misinformation.

\noindent \textbf{\includegraphics[scale=0.1]{Figs/Emoji/pin.pdf} Remarks.}
Temporal misalignment and misinformation pollution are two separate scenarios that give rise to context-memory conflicts. For the former, the up-to-date contextual information is considered accurate. 
\emph{Conversely}, for the latter, the contextual information contains misinformation and is therefore considered incorrect.

\subsection{Analysis of Model Behaviors}
\label{subsec:cm-analysis}

\emph{How do LLMs navigate context-memory conflicts?}
This section will detail the relevant research, although they present quite different answers. 
Depending on the scenario, we first introduce the Open-domain question answering (ODQA) setup and then focus on general setups.

\noindent\textbf{ODQA.} In earlier ODQA literature, \citet{longpre2021entity} explore how QA models act when the provided contextual information contradicts the learned information.
The authors create an automated framework that identifies QA instances with named entity answers, then substitutes mentions of the entity in the gold document with an alternate entity, thus creating the conflict context. This study reveals a tendency of these models to over-rely on parametric knowledge.
\citet{chen2022rich} revisit this setup while reporting differing observations, they note that models predominantly rely on contextual knowledge in their best-performing settings. They attribute this divergence in findings to two factors. Firstly, the entity substitution approach used by \citet{longpre2021entity} potentially reduces the semantic coherence of the perturbed passages. Secondly, \citet{longpre2021entity} based their research on single evidence passages, as opposed to~\citet{chen2022rich}, who utilize multiple ones.
Recently, with the emergence of really ``large'' language models such as ChatGPT~\citep{ouyang2022training, openai2023chatgpt} and Llama 2~\citep{touvron2023llama}, \etc, researchers re-examined this issue. 
\citet{tan2024blinded} examine how LLMs blend retrieved context with generated knowledge in the ODQA setup, and discover models tend to favor the parametric knowledge, influenced by the greater resemblance of these generated contexts to the input questions and the often incomplete nature of the retrieved information, especially within the scope of conflicting sources.

\noindent\textbf{General.} 
\citet{xie2023adaptive} leverage LLMs to generate conflicting context alongside the memorized knowledge. They find that LLMs are highly receptive to external evidence, even when it conflicts with their parametric, provided that the external knowledge is coherent and convincing. Meanwhile, they also identify a strong confirmation bias~\cite{nickerson1998confirmation} in LLMs, \ie, the models tend to favor information consistent with their internal memory, even when confronted with conflicting external evidence. 
\citet{wang2023resolving} posit that the desired behaviors when an LLM encounters conflicts should be to pinpoint the conflicts and provide distinct answers. While LLMs perform well in identifying the existence of knowledge conflicts, they struggle to determine the specific conflicting segments and produce a response with distinct answers amidst conflicting information.
\citet{ying2023intuitive} analyze the robustness of LLMs under conflicts with a focus on two perspectives: factual robustness (the ability to identify correct facts from prompts or memory) and decision style (categorizing LLMs’ behavior as intuitive, dependent, or rational-based on cognitive theory).
The study finds that LLMs are highly susceptible to misleading prompts, especially in the context of commonsense knowledge. 
\citet{qian2023merge} evaluate the potential interaction between parametric and external knowledge more systematically, cooperating knowledge graph (KG). They reveal that LLMs often deviate from their parametric knowledge when presented with direct conflicts or detailed contextual changes.
\citet{xu2023earth} study how LLMs respond to knowledge conflicts during interactive sessions. Their findings suggest LLMs tend to favor logically structured knowledge, even when it contradicts factual accuracy.

\noindent \textbf{\includegraphics[scale=0.1]{Figs/Emoji/pin.pdf} Remarks.}
\noindent \emph{I. Crafting Conflicting Knowledge.}
Model's behavior under context-memory conflict is analyzed by artificially creating conflicting knowledge, in early years through entity-level substitutions and more recently by employing LLMs to generate semantically coherent conflicts.

\noindent \emph{II. What is the conclusion?} 
No definitive rule exists for whether a model prioritizes contextual or parametric knowledge. Yet, knowledge that is \emph{semantically coherent, logical, and compelling} is typically favored by models over generic conflicting information.

\subsection{Solutions}
\label{subsec:cm-mitigation}

Solutions are organized according to their \textbf{objectives}, \ie, the desired behaviors we expect from an LLM when it encounters conflicts.
Existing strategies can be categorized into the following objectives: \emph{Faithful to context} strategies aim to align with contextual knowledge, focusing on context prioritization. \emph{Discriminating misinformation} strategies encourage skepticism towards dubious context in favor of parametric knowledge. 
\emph{Disentangling sources} strategies treat context and knowledge separately and provide disentangled answers.
\emph{Improving factuality} strategies aim for an integrated response leveraging both context and parametric knowledge towards a more truthful solution.

\noindent\textbf{Faithful to Context.}
\noindent \emph{Fine-tuning.} 
\citet{li2022large} argue that an LLM should prioritize context for task-relevant information and rely on internal knowledge when the context is unrelated. 
They name the two properties controllability and robustness. They introduce Knowledge Aware FineTuning (KAFT) to strengthen the two properties by incorporating counterfactual and irrelevant contexts into standard training datasets.
TrueTeacher~\citep{gekhman2023trueteacher} focuses on improving factual consistency in summarization by annotating model-generated summaries with LLMs. This approach helps in maintaining faithfulness to the context of the original documents, ensuring that generated summaries remain accurate without being misled by irrelevant or incorrect details.
DIAL~\citep{xue2023improving} improves factual consistency in dialogue systems via direct knowledge enhancement and reinforcement learning for factual consistency (RLFC) for aligning responses accurately with provided factual knowledge.

\noindent \emph{Prompting.}
\citet{zhou2023context} explores enhancing LLMs' adherence to context through specialized prompting strategies, specifically opinion-based prompts and counterfactual demonstrations. These techniques are shown to significantly improve LLMs' performance in context-sensitive tasks by ensuring they remain faithful to relevant context, without additional training.

\noindent \emph{Decoding.}
\citet{shi2023trusting} introduce Context-aware Decoding (CAD) to reduce hallucinations by amplifying the difference in output probabilities with and without context, similar to the concept of contrastive decoding~\cite{li2022contrastive}. CAD enhances faithfulness in LLMs by prioritizing relevant context over the model's prior knowledge, especially in tasks with conflicting information.

\noindent \emph{Knowledge Plug-in.}
\citet{lee2022plug} propose Continuously-updated QA (CuQA) for improving LMs' ability to integrate new knowledge. Their approach uses plug-and-play modules to store updated knowledge, ensuring the original model remains unaffected. Unlike traditional continued pre-training or fine-tuning approaches, CuQA can solve knowledge conflicts.

\noindent \emph{Pre-training.}
ICLM~\citep{shi2023context} is a new pre-training method that extends LLMs' ability to handle long and varied contexts across multiple documents. This approach could potentially aid in resolving knowledge conflicts by enabling models to synthesize information from broader contexts, thus improving their understanding and application of relevant knowledge.

\noindent \emph{Predict Fact Validity.}
\citet{zhang2023mitigating} address knowledge conflict by introducing fact duration prediction to identify and discard outdated facts in LLMs. This approach improves model performance on tasks like ODQA by ensuring adherence to up-to-date contextual information.

\noindent\textbf{Discriminating Misinformation (Faithful to Memory).}
\noindent \emph{Prompting.}
To address misinformation pollution, \citet{pan2023risk} propose defense strategies such as misinformation detection and vigilant prompting, aiming to enhance the model's ability to remain faithful to factual, parametric information amidst potential misinformation.
Similarly, \citet{xu2023earth} utilize a system prompt to remind the LLM to be cautious about potential misinformation and to verify its memorized knowledge before responding. This approach aims to enhance the LLM's ability to maintain faithfulness.

\noindent \emph{Query Augmentation.}
\citet{weller2022defending} leverage the redundancy of information in large corpora to defend misinformation pollution. Their method involves query augmentation to find a diverse set of less likely poisoned passages, coupled with a confidence method named Confidence from Answer Redundancy, which compares the predicted answer's consistency across retrieved contexts. This strategy mitigates knowledge conflicts by ensuring the model's faithfulness through the cross-verification of answers from multiple sources.

\noindent \emph{Training Discriminator.}
\citet{hong2023discern} fine-tune a smaller LM as a discriminator and combine prompting techniques to develop the model's ability to discriminate between reliable and unreliable information, helping the model remain faithful when confronted with misleading context.

\noindent\textbf{Disentangling Sources.}
DisentQA~\citep{neeman2022disentqa} trains a model that predicts two types of answers for a given question: one based on contextual knowledge and one on parametric knowledge. 
\citet{wang2023resolving} introduce a method to improve LLMs' handling of knowledge conflicts. Their approach is a three-step process designed to help LLMs detect conflicts, accurately identify the conflicting segments, and generate distinct, informed responses based on the conflicting data, aiming for more precise and nuanced model outputs.

\noindent\textbf{Improving Factuality.}
\citet{zhang2023merging} propose COMBO, a framework that pairs compatible generated and retrieved passages to resolve discrepancies. It uses discriminators trained on silver labels to assess passage compatibility, improving ODQA performance by leveraging both LLM-generated (parametric) and external retrieved knowledge.
\citet{jin2024tug} introduce a contrastive-decoding-based algorithm, namely CD$^{2}$, which maximizes the difference between various logits under knowledge conflicts and calibrates the model’s confidence in the truthful answer.

\noindent \textbf{\includegraphics[scale=0.1]{Figs/Emoji/pin.pdf} Remarks.}
Current mitigation approaches have contradicted goals because they do not distinguish between the two causes of knowledge conflict when considering conflict scenarios. Blindly being ``faithful'' to context or knowledge is undesirable.
Some researchers regard that LLM should not rely solely on either parametric or contextual information but instead grant LLM users the agency to make informed decisions based on distinct answers~\citep{wang2023resolving, floridi2023ai}.

\section{\textcolor{lightyellow}{I}nter-\textcolor{lightyellow}{C}ontext Conflict}
\label{sec:inter-context}

Inter-context conflicts manifest in LLMs when incorporating external information sources, a challenge accentuated by the advent of RAG techniques. RAG enriches the LLM's responses by integrating content from retrieved documents into the context. Nonetheless, this incorporation can lead to inconsistencies within the provided context, as the external documents may contain information that conflicts with each other~\cite{zhang2021situatedqa, kasai2022realtime, li2023contradoc}.

\subsection{Causes}
\label{subsec:ic-causes}
\noindent \textbf{Misinformation.} 
Misinformation has long been a significant concern in the modern digital age~\cite{shu2017fake, zubiaga2018detection, kumar2018false, meel2020fake, fung2022battlefront, wang-etal-2023-exploiting}.
The emergence of RAG incorporates external documents to enhance the generation quality of LLMs. While RAG has the potential to enrich content with diverse knowledge sources, it also poses the risk of including documents containing misinformation, such as fake news~\cite{chen2023benchmarking}. 
Moreover, there have been instances where AI technologies are employed to create or propagate misinformation~\cite{weidinger2021ethical, zhou2023synthetic, vergho2024comparing}. The advanced generative capabilities of LLMs exacerbate this issue, leading to an increase in misinformation generated by these systems. This trend is concerning, as it not only contributes to the spread of false information but also challenges detecting misinformation generated by LLMs~\cite{chen2023combating, menczer2023addressing, barrett2023identifying, bengio2023managing, rfid, solaiman2023evaluating,weidinger2023sociotechnical, ferrara2023genai, goldstein2023generative}.

\noindent \textbf{Outdated Information.} 
In addition to the challenge of misinformation, it is important to recognize that facts can evolve. The retrieved documents may contain updated and outdated information from the network simultaneously, leading to conflicts between these documents~\cite{chen2021dataset, liska2022streamingqa, zhang2021situatedqa, kasai2022realtime, SchlichtkrullG023}.  

\noindent \textbf{\includegraphics[scale=0.1]{Figs/Emoji/pin.pdf} Remarks.}
Conflicts in context frequently arise between misinformation and accurate information, as well as between outdated and updated information. These two conflicts exert distinct impacts on LLMs and require specified analysis.
Distinguishing from misinformation conflicts, another significant challenge involves addressing conflicts that arise from documents bearing different timestamps, especially when a user's prompt specifies a particular time period.

\subsection{Analysis of Model Behaviors}
\label{subsec:ic-analysis}
\noindent \textbf{Performance Impact.} 
Previous research empirically demonstrates that the performance of a pre-trained language model can be significantly influenced by the presence of misinformation~\cite{zhang2021situatedqa} or outdated information~\cite{du2022synthetic} within a specific context. In recent studies, \citet{pan2023attacking} introduce a misinformation attack strategy involving the creation of a fabricated version of Wikipedia articles, which is subsequently inserted into the authentic Wikipedia corpus. Their research findings reveal that existing language models are susceptible to misinformation attacks, irrespective of whether the fake articles are manually crafted or generated by models. To gain a deeper understanding of how LLMs behave when encountering contradictory contexts, \citet{chen2022rich} primarily conduct experiments using Fusion-in-Decoder on the NQ-Open~\cite{kwiatkowski2019natural} and TriviaQA~\cite{joshi2017triviaqa}. They find that inconsistencies across knowledge sources exert a minimal effect on the confidence levels of models. These models tend to favor context directly pertinent to the query and context that aligns with the model's inherent parametric knowledge. \citet{xie2023adaptive} conduct experiments on both closed-source LLMs and open-source LLMs in \textsc{PopQA}~\cite{mallen2022not} and \textsc{StrategyQA}~\cite{geva2021did}. The results obtained are in line with those of \citet{chen2022rich}, indicating that LLMs exhibit a significant bias to evidence that aligns with the model's parametric memory. They also find that LLMs exhibit a predisposition towards emphasizing information related to entities of higher popularity and answers that are corroborated by a larger volume of documents within the given context. Moreover, these models demonstrate a significant sensitivity to the order in which data is introduced. \citet{jin2024tug} discover that as the number of conflicting hops increases, LLMs encounter increased challenges in reasoning.

\noindent \textbf{Detection Ability.} 
In addition to assessing the performance of LLMs when confronted with contradictory contexts, several studies also investigate their capacity to identify such contradictions. \citet{zheng2022cdconv} examine the performance of various models including BERT, RoBERTa, and ERNIE in detecting the contradiction within Chinese conversations. Experiments reveal that identifying contradictory statements within a conversation is a significant challenge for these models. \citet{li2023contradoc}
analyse the performance of GPT-4, PaLM-2, and Llama 2 in identifying contradictory documents within news articles~\cite{hermann2015teaching}, stories~\cite{kovcisky2018narrativeqa}, and wikipedia~\cite{merity2016pointer}. The authors find that the average detection accuracy is subpar. The study also finds that LLMs face specific challenges when addressing certain types of contradictions, particularly those involving subjective emotions or perspectives. Additionally, the length of documents and the variety of self-contradictions have a minor influence on the detection performance.
\citet{wan2024evidence} investigate the text features that affect LLMs' assessment of document credibility when faced with conflicting information. They discover that existing models heavily prioritize the relevance of a document to the query but often overlook stylistic features that humans consider important, such as the presence of scientific references or a neutral tone in the text. 
\citet{jin2024tug} discover that LLMs encounter difficulty in distinguishing truthful information from misinformation. In addition, they find that LLMs favor evidence that appears most frequently within the context, and demonstrate confirmation bias for external information aligning with their internal memory.

\noindent \textbf{\includegraphics[scale=0.1]{Figs/Emoji/pin.pdf} Remarks.}
When encountering conflict within a given context, the exhibited knowledge of the LLMs is significantly influenced. However, determining how the model responds to various contextual nuances remains an area requiring further exploration. While different models may share certain commonalities, disparities in behavior arise due to variations in their training data. Moreover, as the model's knowledge is derived from textual information, its approach to discerning misinformation differs significantly from that of humans.

\subsection{Solutions}
\label{subsec:ic-mitigation}
\noindent\textbf{Eliminating Conflict.} 
\noindent \emph{Specialized Models.}
\citet{hsu2021wikicontradiction} develop a model named Pairwise Contradiction Neural Network (PCNN), leveraging fine-tuned Sentence-BERT embeddings to calculate contradiction probabilities of articles.  \citet{pielka2022linguistic} suggest incorporating linguistic knowledge into the learning process based on the discovery that XLM-RoBERTa struggles to effectively grasp the syntactic and semantic features that are vital for accurate contradiction detection. \citet{wu2022topological} propose an innovative approach that integrates topological representations of text into language models to enhance the contradiction detection ability and evaluated their methods on the MultiNLI dataset~\cite{williams2017broad}.

\noindent \emph{General Models.}
\citet{chern2023factool} propose a fact-checking framework that integrates LLMs with various tools, including Google Search, Google Scholar, code interpreters, and Python, for detecting factual errors in texts. \citet{leite2023detecting} employ LLMs to generate weak labels associated with predefined credibility signals for the input text and aggregate these labels through weak supervision techniques to make predictions regarding the veracity of the input. 

\noindent\textbf{Improving Robustness.} 
\noindent \emph{Training Approach.}
\citet{hong2023discern} present a novel fine-tuning method that involves training a discriminator and a decoder simultaneously using a shared encoder. Additionally, the authors introduce two other strategies to improve the robustness of the model including prompting GPT-3 to identify perturbed documents before generating responses and integrating the discriminator's output into the prompt for GPT-3. Their experimental results indicate that the fine-tuning method yields the most promising results.

\noindent \emph{Query Augmentation.}
\citet{weller2022defending} explore a query augmentation technique that prompts GPT-3 to formulate new questions derived from the original inquiry. They then assess the confidence for each answer by referencing the corresponding passages retrieved. Based on the confidence, they decide whether to rely on the original question's prediction or aggregate predictions from the augmented questions with high confidence scores.

\noindent \textbf{\includegraphics[scale=0.1]{Figs/Emoji/pin.pdf} Remarks.}
Strategies for addressing inter-context conflicts primarily rely on model knowledge or leveraging external knowledge such as retrieved documents. Recently, augmenting LLM with external tools has emerged as a new paradigm. Exploring the use of external tools to support LLMs in resolving inter-context conflicts could be a promising approach. On the other hand, devising unified and efficient methods to handle various conflict types remains a formidable challenge.

\section{\textcolor{cyan}{I}ntra-\textcolor{cyan}{M}emory Conflict}
\label{sec:intra-memory}
With the development of LLMs, LLMs are widely used in knowledge-intensive question-and-answer systems~\cite{gao2023retrieval, yu2022generate, petroni2019language, chen2023beyond}. A critical aspect of deploying LLMs effectively involves ensuring that they produce consistent outputs across various expressions that share similar meanings or intentions. 
Despite this necessity, a notable challenge arises with intra-memory conflict—a condition where LLMs exhibit unpredictable behaviors and generate differing responses to inputs that are semantically equivalent but syntactically distinct~\cite{chang2023language, chen2023say, raj2023semantic, rabinovich2023predicting, raj2022measuring, bartsch2023self}. Intra-memory conflict essentially undermines the reliability and utility of LLMs by introducing a degree of uncertainty in their output.

\subsection{Causes}
\label{subsec:im-causes}
Intra-memory conflicts within LLMs can be attributed to three primary factors: training corpus bias~\cite{wang2023causal,xu2022does}, decoding strategies~\citet{lee2022factuality,huang2023survey}, and knowledge editing~\cite{yao2023editing,li2023unveiling}. These factors respectively pertain to the training phase, the inference phase, and subsequent knowledge refinement. 

\noindent \textbf{Bias in Training Corpora.}
Recent research demonstrates that the primary phase for knowledge acquisition in LLMs predominantly occurs in the pre-training stage~\cite{zhou2023lima, kaddour2023challenges, naveed2023comprehensive, akyurek2022towards, singhal2022large}. Pre-training corpus is primarily crawled from the internet, which exhibits a diverse range of data quality, potentially including inaccurate or misleading information~\cite{bender2021dangers, weidinger2021ethical}. When LLMs are trained on data containing incorrect knowledge, they may memorize and inadvertently amplify these inaccuracies~\cite{lin2022truthfulqa, elazar2022measuring, lam2022analyzing, grosse2023studying}, leading to a situation where conflicting knowledge coexists within the parameters of LLMs. 

Moreover, prior works indicate that LLMs tend to encode superficial associations prevalent within their training data, as opposed to genuinely comprehending the underlying knowledge contained therein~\cite{li2022pre, kang2023impact, zhao2023explainability, kandpal2023large}. It can result in LLMs displaying a propensity to generate predetermined responses rooted in spurious correlations of training data. Due to the dependency on spurious correlations, LLMs may provide divergent answers when presented with prompts exhibiting distinct syntactic structures but conveying equivalent semantic meaning.

\noindent \textbf{Decoding Strategy.}
The direct output of LLMs is a probability distribution over potential next tokens. Sampling is a crucial step in determining the generated content from this distribution. Various sampling techniques, including greedy sampling, top-p sampling, top-k sampling, and others have been proposed~\cite{jawahar2020automatic, massarelli2019decoding}, broadly categorizing into deterministic and stochastic sampling methods. Stochastic sampling stands as the prevailing decoding strategy employed by LLMs~\cite{fan2018hierarchical, holtzman2019curious}. 
However, the random nature of stochastic sampling methods introduces uncertainty into the generated content. Moreover, due to the intrinsic left-to-right generation pattern inherent to LLMs, the selection of the sampling token can wield a significant influence over the subsequent generations. The use of stochastic sampling may cause LLMs to produce entirely different content, even when provided with the same context, causing intra-memory conflict~\cite{lee2022factuality,huang2023survey, dziri2021neural}.

\noindent \textbf{Knowledge Editing.}
With the exponential increase of model parameters, fine-tuning LLMs become increasingly challenging and resource-intensive. In response to this challenge, researchers explore knowledge editing techniques as a means of efficiently modify a small scope of the knowledge encoded in LLMs~\cite{meng2022locating, ilharco2022editing, zhong2023mquake}. 
Ensuring the consistency of modifications poses a significant challenge. Due to the potential limitations inherent in the editing method, the modified knowledge cannot be generalized effectively. This can result in LLMs producing inconsistent responses when dealing with the same piece of knowledge in varying situations~\cite{li2023unveiling, yao2023editing}. Intra-memory conflict is primarily considered a side effect in the context of knowledge editing.

\noindent \textbf{\includegraphics[scale=0.1]{Figs/Emoji/pin.pdf} Remarks.}
Intra-memory conflicts in LLMs arise from three distinct causes that occur at different stages. Among these causes, training corpus bias stands out as the fundamental catalyst. Incongruities of knowledge in
the training dataset result in inconsistencies within the knowledge encoded within the model's parameters. Additionally, the decoding strategy indirectly contributes to exacerbating these conflicts. The inherent randomness of the sampling process during inference amplifies the inconsistencies in the model's responses. Knowledge editing, which aims to post-update the model's knowledge, can inadvertently introduce conflicting information into the LLM's memory.

\subsection{Analysis of Model Behaviors}
\label{subsec:im-analysis}
\noindent \textbf{Self-Inconsistency.}
\citet{elazar2021measuring} develop a method for assessing the knowledge consistency of language models, focusing specifically on knowledge triples. The authors primarily conduct experiments using BERT, RoBERTa, and ALBERT. Their findings indicate that these models exhibit poor consistency, with accuracy rates barely ranging from 50\% to 60\%. \citet{hase2023methods} employ the same indicators of \citet{elazar2021measuring}, but they utilize a more diverse dataset. Their study also reveals that the consistency of RoBERTa-base and BART-base within the paraphrase context is lacking. \citet{zhao2023knowing} reformulate questions and then assess the consistency of the LLM's responses to these reformulated questions. The findings of their research reveal that even GPT-4 exhibits a notable inconsistency rate of 13\% when applied to Commonsense Question-Answering tasks. They further find that LLMs are more likely to produce inconsistencies in the face of uncommon knowledge. \citet{dong2023statistical} conduct experiments on multiple open-source LLMs and find that all of these models exhibit strong inconsistencies. \citet{li2023benchmarking} explore an additional aspect of inconsistency that LLMs can give an initial answer to a question, but it may subsequently deny the previous answer when asked if it is correct. The authors conduct experiments focusing on Close-Book Question Answering and reveal that Alpaca-30B only displays consistency in 50\% of cases.

To further analyze the inconsistency exhibited by LLMs, a study conducted by \citet{li2022pre} reveals that encoder-based models tend to generate mis-factual words more relying on positionally close and highly co-occurring words, rather than knowledge-dependent words. This phenomenon arises due to these models' tendency to overlearn inappropriate associations from the training dataset. \citet{kang2023impact} highlight a co-occurrence bias in LLMs, where the models favor frequently co-occurring words over correct answers. especially when recalling facts where the subject and object rarely co-occur in the pre-training dataset, despite fine-tuning. Furthermore, their research indicates that LLMs face challenges in recalling facts in cases where the subject and object rarely appear together in the pre-training dataset, even though these facts are encountered during fine-tuning.

\noindent \textbf{Latent Representation of Knowledge.}
The multi-layer transformer architecture inherent to contemporary LLMs fosters a complex inter-memory conflict, with distinct knowledge representations scattered across various layers. Previous research suggests that LLMs store low-level information at shallower levels and semantic information at deeper levels~\cite{tenney2019bert, rogers2021primer, wang2019make, jawahar2019does, cui2020does}. \citet{chuang2023dola} explore this aspect within the context of LLMs and discover that the factual knowledge in LLMs is typically concentrated within specific transformer layers and different layers of inconsistent knowledge. Moreover, \citet{li2023inference} discover that the correct knowledge is indeed stored within the parameters of the model, but it may not be accurately expressed during the generation process. The authors conduct two experiments on the same LLM, one focused on the generation accuracy, and the other utilizing a knowledge probe to examine the knowledge containment. The results of these experiments reveal a substantial 40\% disparity between the knowledge probe accuracy and the generation accuracy. 

\noindent \textbf{Cross-lingual Inconsistency.}
The universality of true knowledge transcends surface form variations~\cite{ohmer2023separating}, a principle that should ideally apply to LLMs. However, LLMs maintain distinct knowledge sets for different languages, leading to inconsistencies~\cite{ji2023survey, xue2024comprehensive}. \citet{wang2023cross} investigate the challenges LLMs face in extending edited knowledge across languages, suggesting that knowledge related to different languages is stored separately within the model parameters. \citet{qi2023cross} propose a metric named RankC for evaluating the cross-lingual consistency of LLMs' factual knowledge. They employ this metric for analyzing multiple models and reveal a pronounced language dependence in the knowledge stored by LLMs, with no observed improvement in cross-lingual consistency with increased model size.

\noindent \textbf{\includegraphics[scale=0.1]{Figs/Emoji/pin.pdf} Remarks.}
The phenomenon of inter-memory conflict in LLMs predominantly manifests through inconsistent responses to semantically identical queries. This inconsistency is primarily attributed to the suboptimal quality of datasets utilized during the pre-training phase. Addressing this challenge necessitates the development of efficient and cost-effective solutions, which remains a significant hurdle. Additionally, LLMs are characterized by the presence of multiple knowledge circuits, which significantly influence their response mechanisms to specific inquiries. The exploration and detailed examination of these knowledge circuits within LLMs represent a promising avenue for future research.

\subsection{Solutions}
\label{subsec:im-mitigation}
\subsubsection{Improving Consistency}

\noindent \emph{Fine-tuning.}
\citet{elazar2021measuring} propose a consistency loss function and train the language model with the combination of the consistency loss and standard MLM loss. 
\citet{li2023benchmarking} utilize one language model in dual capacities: as a generator to produce responses and as a validator to evaluate the accuracy of these responses. The process involves querying the generator for a response, which is subsequently assessed by the validator for accuracy. Only those pairs of responses deemed consistent are retained. This subset of consistent pairs is then used to fine-tune the model, aiming to increase the generation likelihood of consistent response pairs.

\noindent \emph{Plug-in.}
\citet{jang2023improving} leverage the technique of intermediate training, utilizing word-definition pairs from dictionaries to retrain language models and improve their comprehension of symbolic meanings. Subsequently, they propose an efficient parameter integration approach, which amalgamates these enhanced parameters with those of existing language models. This method aims to rectify the models' inconsistent behavior by bolstering their capacity to understand meanings.

\noindent \emph{Output Ensemble.}
\citet{mitchell2022enhancing} mitigate the inconsistency of language models by leveraging a two-model architecture, involving the utilization of a base model responsible for generating a set of potential answers, followed by a relation model that evaluates the logical coherence among these answers. The final answer is selected by considering both the base model's and the relation model's beliefs. \citet{zhao2023knowing} introduce a method to detect whether a question may cause inconsistency for LLMs. Specifically, they first use LLMs to rephrase the original question and obtain corresponding answers. They then cluster these answers and examine the divergence. The detection is determined based on the divergence level.

\subsubsection{Improving Factuality}
\citet{chuang2023dola} propose a novel contrastive decoding approach named DoLa. Specifically, the authors develop a dynamic layer selection strategy, choosing the appropriate premature layers and mature layers. The next word's output probability is then determined by computing the difference in log probabilities of the premature layers and the mature layers. 
\citet{li2023inference} propose a similar method named ITI. They first identify a sparse set of attention heads that exhibit high linear probing accuracy for truthfulness, as measured by TruthfulQA~\cite{lin2022truthfulqa}. During the inference phase, ITI shifts activations along the truth-correlated direction, which is obtained through knowledge probing. This intervention is repeated autoregressively for every token during completion. Both DoLa and ITI address the inconsistency of knowledge across the model's different layers to reduce factual errors in LLMs.

\noindent \textbf{\includegraphics[scale=0.1]{Figs/Emoji/pin.pdf} Remarks.}
The resolution of inter-memory conflict in LLMs typically entails three phases: training, generation, and post-hoc processing. The training phase method mainly focuses on mitigating internal inconsistencies among model parameters. Conversely, the generation and post-hoc phases primarily involve algorithmic interventions aimed at alleviating occurrences of inconsistent model behavior. Nevertheless, the challenge persists in addressing the inconsistency of parameter knowledge without detrimentally impacting the overall performance of LLMs.

\section{Challenges and Future Directions}
\label{sec:challenges}

In this section, we provide a summary and highlight the existing challenges in ongoing research, as well as outline potential future directions in the field of knowledge conflict.

\noindent\textbf{Knowledge Conflicts in the Wild.} 
Currently, the creation of knowledge conflicts predominantly relies on the artificial generation of incorrect or misleading information. In the real world, one of the most common situations where knowledge conflicts arise is in RALMs (Retrieval-Augmented Language Models), where conflicts are present in documents retrieved by the retrieval module directly from the Web. Current analysis approaches exist a gap in the experimental setup of knowledge conflict, suggesting that findings from those environments~\cite{xie2023adaptive,wang2023resolving} might not easily transfer to practical applications. 
Recent studies have begun to investigate the scenario in the wild by curating conflicting documents based on actual search results from Google for open-ended questions~\cite{wan2024evidence}. 
Looking ahead, there is a growing interest in more research that assesses how well LLMs perform in real-world scenarios rather than artificially created conflicts, to better understand their capabilities.

\noindent\textbf{Solution at a Finer Resolution.}
Currently, there is no one-size-fits-all solution to knowledge conflict due to its inherent complexity. Existing either assume a prior~\cite{shi2023context} or focus on a subclass of conflict~\cite{wang2023resolving}.
We believe that addressing this issue requires a more fine-grained approach, taking into account several factors. Firstly, the nature of the user's query plays a crucial role. Subjective or debatable questions naturally lead to conflicts as they may have multiple valid answers~\cite{bjerva2020subjqa,wan2024evidence}. 
Secondly, the source of conflicting information can vary, including misinformation, outdated facts, or partially correct data~\cite{guo2022survey, AkhtarSGCS023}. 
Lastly, it is also important to consider user expectations, such as whether they prefer a single definitive answer from the LLM or are open to multiple perspectives~\cite{floridi2023ai}.
Given these considerations, future solutions to mitigate knowledge conflicts must delve into these nuances, recognizing that knowledge conflict encompasses a spectrum of problems with diverse causes, manifestations, and potential resolutions. 
Collaboration between NLP and HCI researchers is appreciated for conducting thorough investigations and developing effective solutions.

\noindent\textbf{Evaluation on Downstream Tasks.}
The current landscape of research into knowledge conflicts within LLMs predominantly emphasizes evaluating their performance on common QA datasets including NQ-Open, TriviaQA, OPQA, and \textsc{StrategyQA}. This focus overlooks the broader implications of knowledge conflicts, particularly how they influence downstream tasks. Exploring the effects of knowledge conflicts on a wider range of applications beyond QA problems could yield insights into creating more robust and reliable models. For instance, in tasks requiring high levels of accuracy and consistency, such as legal document analysis~\cite{shui2023comprehensive, martin2024better}, medical diagnosis~\cite{zhou2023survey, thirunavukarasu2023large}, financial analysis~\cite{zhang2023enhancing, li2023large} and educational tools~\cite{caines2023application, milano2023large}, the presence of unresolved knowledge conflicts could undermine the model's utility.

\noindent\textbf{Interplay among the Conflicts.}
Current research in the knowledge conflict of LLMs primarily concentrates on investigating conflicts of a singular type ~\cite{wang2023resolving, chen2022rich, li2023benchmarking} or a joint study of inter-context and context-memory conflict~\cite{jin2024tug, xie2023adaptive}. However, there is a notable dearth of research on the interaction between intra-memory conflict and the other two types of conflicts. Several papers have proposed the existence of knowledge circuits in LLMs~\cite{chughtai2024summing, huang2023survey}, which are closely related to the intra-memory conflict.
Addressing this gap is crucial for understanding the relationship between the internal knowledge inconsistency of the model and its behavior in response to the context. Moreover, exploring the synergistic effects of various conflict types could unveil underlying mechanisms of knowledge representation and processing in LLMs and help us to develop more robust and accurate LLMs in practice.

\noindent\textbf{Explainability.}
Most recent work analyzed LLMs' behaviors amidst knowledge conflicts at the output level~\cite{xie2023adaptive,wang2023resolving}. While some studies have observed and explored the model's confidence in its output, \ie, logits~\cite{xu2023earth,jin2024tug,wang2024selfdc}, there has been less focus on the internal mechanism of the model, like specific attention heads or neuron activations during conflicts. This gap highlights a need for more microscopic examinations to better comprehend how models decide when encounter conflicts. 
A recent study conducted by \citet{jin2024cutting} advances this by investigating the interpretability of LLMs through information flow analysis, pinpointing pivotal points for conflict mitigation. They discover there are some attention heads with opposite effects in the later layers, where memory heads can recall knowledge from internal memory, and context heads can retrieve knowledge from the external context. Inspired by this, Pruning Head via Path Patching is introduced to resolve conflicts efficiently without updating model parameters.

\noindent\textbf{Multilinguality.}
To date, research on knowledge conflict has primarily focused on the English language. Future studies could expand in two directions. 
Firstly, by examining LLMs to address knowledge conflicts in non-English prompts, leveraging the many advanced non-English models available (\eg, GLM~\cite{zeng2022glm} for Chinese) or LLMs with multilingual capability (\eg, GPT-4~\cite{openai2023gpt4}) and noting differences from English to account for unique language characteristics. 
Secondly, addressing inter-context conflict where multiple documents in different languages might be retrieved, potentially involving cross-language knowledge conflicts. Solutions could include employing translation systems~\cite{dementieva2021cross} or, for low-resource languages, leveraging high-resource language evidence \citep{xue2024comprehensive} or employing knowledge distillation techniques.

\noindent\textbf{Multimodality.}
Current research on knowledge conflicts mainly focused on the text modality, leaving the study of these conflicts in multimodal contexts as a promising area for future exploration.
As LLMs evolve to process information across various formats---images~\cite{alayrac2022flamingo,li2023blip}, video~\cite{ju2022prompting,zhang2023video}, and audio~\cite{borsos2023audiolm,wu2023large}---the potential for conflicts escalates in growing complexity.
For instance, textual documents might clash with visual data, or the tone of an audio clip might contradict the content of an accompanying caption.
Future research on multimodal knowledge conflicts could focus on crafting advanced LLMs skilled in cross-modal reasoning and conflict resolution across diverse data types. This effort necessitates the enhancement of models' capabilities to navigate the complex dynamics between different modalities and the development of targeted datasets for effective training and evaluation. Additionally, exploring how users perceive and manage multimodal conflicts, such as discrepancies between text and images, will offer valuable insights into improving LLMs for better human interaction.

\section{Conclusion}
\label{sec:conclusion}

Through this survey, we have extensively investigated knowledge conflicts, shedding light on their categorization, causes, how LLMs respond to these conflicts, and possible solutions. 
Our findings reveal that knowledge conflict is a multifaceted issue, with a model's behavior being closely tied to the particular type of conflicting knowledge. Besides, there appears to be a more complex interplay among the three types of conflicts. 
Furthermore, we observe that existing solutions primarily address artificially constructed scenarios, neglecting the subtleties of conflicts by relying on assumed priors and thus sacrificing granularity and breadth.
Given the growing use of retrieval-augmented language models, we anticipate that knowledge conflicts faced by LLMs will only increase in complexity, underscoring the need for more comprehensive research.

\section*{Limitations}
Considering the rapid expansion of research in the field of knowledge conflict and the abundance of scholarly literature, it is possible that we might have missed some of the most recent or less relevant findings. Nevertheless, we have ensured the inclusion of all essential materials in our survey.

\section*{Ethics Statement}
We mainly searched for papers published after 2021 using key terms including ``knowledge conflict'', ``knowledge inconsistency'', ``knowledge gap'', \etc{}, on Google Scholar and the ACL Anthology. After initially identifying these papers, the authors classified them through reading and continued to track related but overlooked papers using their citations. We also used Google Scholar to follow up on the latest papers citing these to avoid omissions. 

For the quantitative analysis and comparison section (\autoref{sec: quantitative}), we did not conduct computational experiments but simply organized the result reported in other literature as is.

\bibliography{anthology,custom}
\bibliographystyle{acl_natbib}

\appendix
\newpage

\begin{table*}[h]
\fontsize{10}{10}\selectfont
\centering
\setlength{\tabcolsep}{2pt} %
\begin{threeparttable}
\begin{tabularx}{\textwidth}{>
{\centering\arraybackslash}p{2.5cm}>{\centering\arraybackslash}p{2.5cm}>{\centering\arraybackslash}p{3.35cm}>{\centering\arraybackslash}X}
\toprule
\textbf{Reference} & \textbf{Model} & \textbf{Dataset} & \textbf{Quantitative Results}\\
\midrule
\multicolumn{4}{c}{\emph{Context-memory conflict}}\\
\midrule
\citet{pan2023risk} & ChatGPT & NQ-1500 and CovidNews & Misinformation in the context can lead to a significant degradation (up to 87\%) in the performance.\\
\midrule
\citet{xie2023adaptive} & ChatGPT, GPT-4, PaLM2, Qwen, Llama2, and Vicuna & POPQA and STRATEGYQA & For entity substitution-based counter-memory, only ChatGPT, GPT-4, and PaLM2 over 60\% probability of choosing parametric memory. For generation-based counter-memory, all models have more than 80\% probability of choosing context knowledge.\\
\midrule
\citet{xu2023earth} & ChatGPT, GPT-4, Llama2, and Vicuna & Farm, BoolQ, TruthfulQA and NQ & In multiple rounds of dialogue, as the number of counter-memory context increases, the cumulative proportion of belief alteration of LLMs spans from 20.7\% to 78.2\%\\
\midrule
\multicolumn{4}{c}{\emph{Inter-context conflict}}\\
\midrule
\citet{jin2024tug} & ChatGPT, Llama2, Baichuan2, FLAN-UL2 and FLAN-T5 &  NQ, TriviaQA, PopQA, and MuSiQue & When faced with conflicting evidence, ChatGPT's recall declined the least, but more than 10\%.\\
\midrule
\citet{chen2023benchmarking} & ChatGPT, ChatGLM, Vicuna, Qwen, and BELLE & RGB & As the noise in evidence increases, the performance of models will gradually decrease. When the noise rate exceeds 0.8, the performance of all models decreases by more than 20\%.\\
\midrule
\citet{li2023contradoc} & GPT-4, ChatGPT, PaLM2, and Llama2 & CONTRADOC & Faced with self-contradictory documents, gpt4 has a more than 70\% probability of determining the occurrence of a contradiction, while other models are less than 50\%. \\
\midrule
\multicolumn{4}{c}{\emph{Intra-memory conflict}}\\
\midrule
\citet{mundler2023self} & 
GPT-4, ChatGPT, Llama2, and Vicuna & MainTestSet & LLMs create contradictory content, with a probability of between 15.7\% and 22.9\%. More powerful models create fewer contradictory results. \\
\midrule
\citet{zhao2023knowing} & ChatGPT, GPT-4, Vicuna, and Llama2 & FaVIQ, ComQA, GSM-8K, SVAMP, ARCChallenge, and CommonsenseQA & The findings of their research reveal that even GPT-4 can exhibit an inconsistency rate of 32\% in FaVIQ. \\
\bottomrule
\end{tabularx}
\end{threeparttable}
\caption{\label{tab: quantitative_analysis}
Comparison of quantitative results on the impact of various types of knowledge conflicts.
}
\end{table*}

\section{Appendix}
\label{sec:appendix}

\subsection{Quantitative Analysis and Comparison}
\label{sec: quantitative}

In the context of a survey paper, while it is beneficial to include quantitative results and analyses concerning the impact of knowledge conflicts across various types of conflicts and the performance comparison of different mitigation strategies, it is not a strict requirement. 
We acknowledge the \emph{complexity and impracticality} involved in conducting such quantitative experiments, particularly due to the use of disparate datasets in behavioral analyses, as well as the variance in the inherent knowledge of LLMs across different knowledge cut-off snapshots, as detailed in~\autoref{tab:datasets}.

Moreover, establishing a ``fair'' comparison within the mitigation strategies segment poses its own set of challenges, given the diversity in objectives influenced by various assumed priors, such as the perceived accuracy of context or inherent knowledge, as discussed in the main text.
Despite these intricacies, we opt to present quantitative results by compiling existing evaluations from a range of papers. \emph{It is imperative, however, to \textbf{approach this analysis with caution}, recognizing that original authors may have employed different datasets, LLMs variants, or even pursued contrasting objectives.}

\subsection{Quantitative Results on the Impact of Knowledge Conflicts}

The comparison of quantitative results on the impact of the three types of knowledge conflicts is shown in~\autoref{tab: quantitative_analysis}. We pick the results of representative behavior analysis literature for comparison.

\subsection{Quantitative Results on the Effectiveness of Mitigation Strategies}

\begin{table*}[h]
\fontsize{10}{10}\selectfont
\centering
\setlength{\tabcolsep}{2pt} %
\begin{threeparttable}
\begin{tabularx}{\textwidth}{>
{\centering\arraybackslash}p{2.5cm}>{\centering\arraybackslash}p{2.5cm}>{\centering\arraybackslash}p{3.35cm}>{\centering\arraybackslash}X}
\toprule
\textbf{Reference} & \textbf{Model} & \textbf{Dataset} & \textbf{Quantitative Results}\\
\midrule
\multicolumn{4}{c}{\emph{Faithful to context}}\\
\midrule
\citet{shi2023trusting} & Llama, OPT, GPT-Neo, and FLAN &  NQ-SWAP, MemoTrap, and NQ & Their method improves GPT-Neo 20B by 54.4\% on Memotrap and by 128\% on NQ-SWAP where LLMs need to adhere to the given context. \\
\midrule 
\citet{zhou2023context} & ChatGPT and Llama2 & MRC and Re-TACRED & Compared to the zero-shot base prompts, their prompting method leads to a reduction of 32.2\% for maintaining parametric knowledge for MRC and a 10.9\% reduction
for Re-TACRED on GPT-3.5. Similarly, on Llama2, there is a 39.4\% reduction for MRC and a 57.3\% reduction for Re-TACRED. \\
\midrule
\multicolumn{4}{c}{\emph{Discriminating misinformation}}\\
\midrule
\citet{hong2023discern} & ChatGPT and FiD & NQ and TQA & The authors train a discriminator with about 80\% F1 score and use it to improve models performance above 5\%. \\
\midrule
\citet{pan2023risk} & ChatGPT & NQ-1500 and CovidNews & The author's mitigation method improves the accuracy by more than 10\%.\\
\midrule
\multicolumn{4}{c}{\emph{Disentangling sources}}\\
\midrule
\citet{wang2023resolving} & ChatGPT & KNOWLEDGE CONFLICT & The authors' method achieved over 80\% F1 score on contextual knowledge conflict detection.\\
\bottomrule
\end{tabularx}
\end{threeparttable}
\caption{\label{tab: quantitative_mitigation}
Comparison of quantitative results on the effectiveness of various mitigation strategies \wrt{} their objectives.
}
\end{table*}

The effectiveness of various mitigation strategies is quantitatively compared in~\autoref{tab: quantitative_mitigation}. It is important to note that our analysis is limited to works addressing \emph{three predominant types of mitigating objectives} within the context of memory conflicts. This selection is deliberate, as other types of mitigating objectives in different conflict categories do not yet have a substantial body of work that would allow for a meaningful cross-method comparison.

\end{document}